\documentclass[10pt,twocolumn,letterpaper]{article}
\usepackage[pagenumbers]{aaai} % To force page numbers, e.g. for an arXiv version

\usepackage{graphicx}
\usepackage{amsmath}
\usepackage{amssymb}
\usepackage{booktabs}
\usepackage{subfloat}

\usepackage[pagebackref,breaklinks,colorlinks]{hyperref}

\usepackage[capitalize]{cleveref}

\begin{document}

\title{Machine unlearning via GAN}

\author{Kongyang Chen\\
{\tt\small kongyang.chen@gmail.com}

\and

Yao Huang\\

{\tt\small huangyao0715@gmail.com}

\and

Yiwen Wang\\
{\tt\small 984985386@qq.com}

}

\maketitle

\begin{abstract}
Machine learning models, especially deep models, may unintentionally remember information about their training data. Malicious attackers can thus pilfer some property about training data by attacking the model via membership inference attack or model inversion attack. Some regulations, such as the EU's GDPR, have enacted "The Right to Be Forgotten" to protect users' data privacy, enhancing individuals' sovereignty over their data. Therefore, removing training data information from a trained model has become a critical issue. In this paper, we present a GAN-based algorithm to delete data in deep models, which significantly improves deleting speed compared to retraining from scratch, especially in complicated scenarios. We have experimented on five commonly used datasets, and the experimental results show the efficiency of our method.
\end{abstract}

\section{Introduction}

Many research have shown that machine learning model, especially over-parameterized model, could remember a lot of information about the training data\cite{carlini2019secret}\cite{feldman2020does}, malicious attacker could thus leverage model inversion attack\cite{fredrikson2015model} to recover the model's training data, or utilizing membership inference attack\cite{shokri2017membership}\cite{salem2019ml} to discriminate whether a certain data is used to train the model, which seriously violate user’s data privacy. In order to better protect user's data privacy, recently enacted laws, For instance, GDPR\cite{voigt2017eu} and CCPA\cite{CCPA}, have clearly proposed \textbf{\textit{The Right to be Forgotten}}. It states that data subjects(users) have the right to request data controllers to delete their personal data, and the controller is responsible for deleting personal data in a timely way under certain conditions. When data controllers receive a data detection request from user, they should not only remove data from the hard drive, but also erase any remaining information in trained model. Removing data lineage from trained model is named \textbf{\textit{Mahcine Unleaning}}, It was originally proposed by\cite{cao2015towards}.
A naive method for removing the data lineage from trained model is to retrain model from scratch, because the user's data is never involved in the retraining process, thus retraining could naturally erase the lineage of user’s data. In complex scenarios, however, retraining would bring high computational and time costs. As a result, the most pressing issue in machine unlearning is how to reduce costs while increasing efficiency. \cite{cao2015towards}Proposed statistic query-based unlearning, in which they induced a summation term between model and raw dataset, the model was trained on the summation term rather than raw data, and when deleting data, they simply removed the data from raw dataset and formed a new summation term, then retrain model on the new summation term. It accelerates the unlearning process, but it is not applicable to neural networks. \cite{bourtoule2019machine}partitioned original dataset into disjoint data chunks, each data chunk trains its own model independently, and the final model was integrated by these models, erasing data only need to retrain the corresponding model. To accelerate retraining, they further divided each data chunks into disjoint data slices to save intermedia parameters of model. The disadvantages of this strategy is that when the model is large, storing the model's parameters would consume a lot of storage space. Recently, \cite{liu2020federated}studied removing data from a trained model in federated learning scenario, similar to\cite{bourtoule2019machine}, the retraining process can be accelerated by caching the intermediate parameters of model. \cite{brophy2021machine}studied how to erase data in random forests.

Deep classifiers are widely used in various fields, such as face recognition, object detection, etc. In this paper, we focus on deep classifiers, and we intend to erase data in deep classifiers efficiently while minimizing performance degradation as possible. We proposed a fast data deletion method from a trained model via generative adversarial networks. Our approach does not need to cache any intermedia parameters, and the deleting speed is significantly improved compared to retraining. Moreover, our method is also convenient to deploy. It can apply to most existing machine learning frameworks without any modification. We evaluated our approach on different datasets and different model architectures.

Our contributions are as follows:
\begin{itemize}
    \item We introduced generative adversarial network into Machine unlearning and proposed a method based on GAN to approximately delete data, which is fast, efficient, and easy to deploy. To the best of our knowledge, we are the first to delete data from a trained model via GAN.

    \item Our method does not need to retrain the model. Compared with retraining model from scratch naively, our approach is faster and has a lower computational cost, especially in complex scenarios. 
    
    \item Our method is memory free, which means we don't need any storage resources to cache the intermedia parameters of model. 
\end{itemize}

\section{Method}
\subsection{Overview}
\begin{figure*}[htbp]
    \centering
    \includegraphics[scale=0.5]{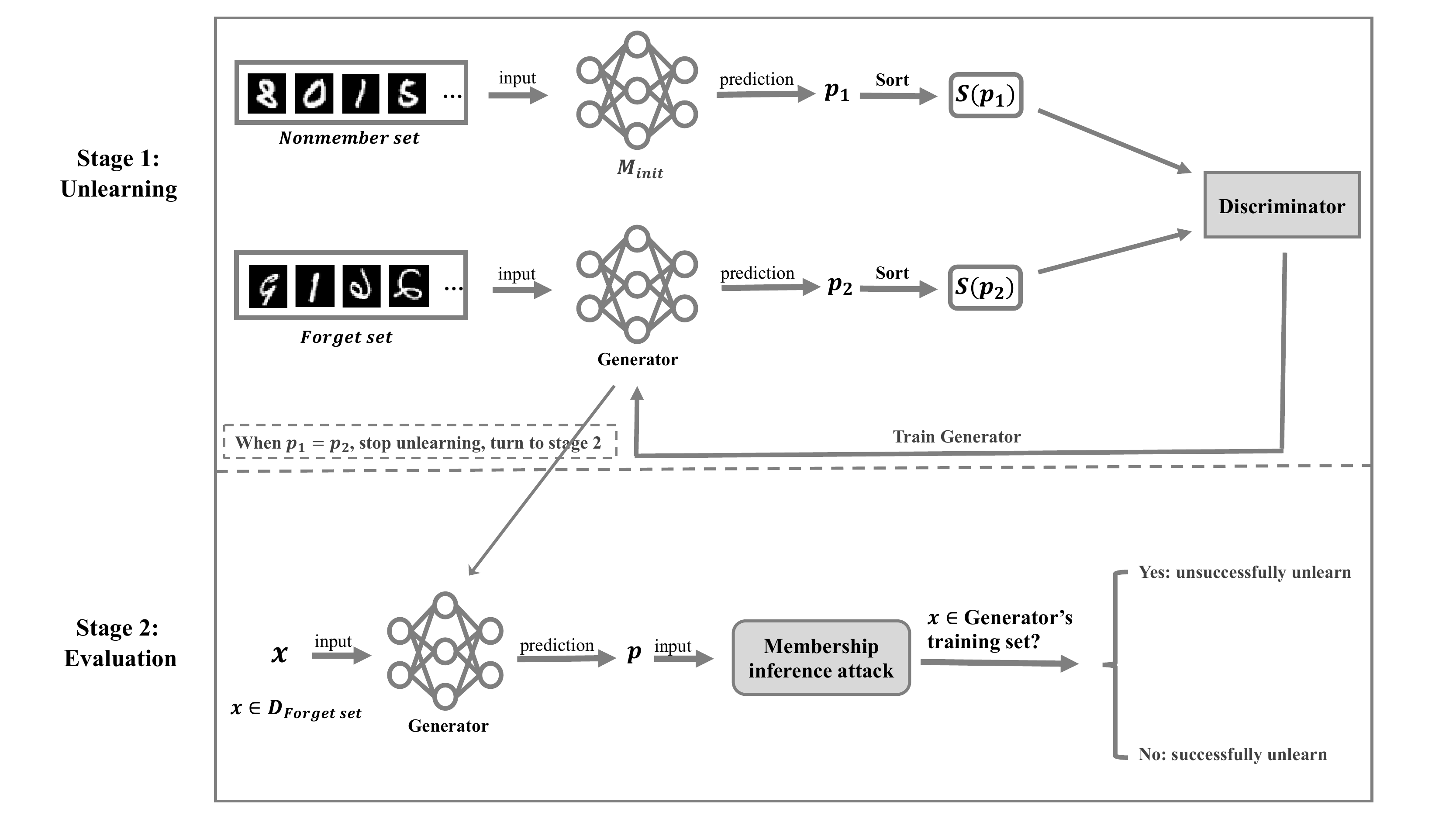}
    \caption{framwork.}
    \label{fig:framework}
\end{figure*}

Assuming that we are an Internet service provider, such as Google, Facebook, Amazon, etc., the service provider collects user data and trains the model based on collected data, then deploys the trained model as a service. If users request to delete data, the service provider needs to delete the user's data both from the trained model and hard drive under the corresponding laws (like GDPR).We assume the data needs to be deleted is $D_f$, and the trained model is $M_{init}$. Our goal is to delete $D_f$ from $M_{init}$.To fast and cost-effectively delete data, we treat the data deletion issue as making $D_f$ behave similar to the third-party data in the model. After deleting data $D_f$, the posterior distribution of $D_f$ and third-party should be the same. Here, The third-party data means that it has the same distribution as $D_f$ but never trained the model.To achieve the above goal, we proposed Fast Data Forgotten based on GAN. The generator of GAN is initialized as the trained model $M_{init}$, then the generator and discriminator are trained alternately until the discriminator cannot distinguish the output distribution difference of model between $D_f$ and third-party data. Moreover, in order to reduce performance degradation, we modified the loss function of the generator by introducing the hyperparameter $\alpha$. Our approach achieved a balance between deletion and performance. To evaluate the effect of our approach, we conducted experiments from three aspects: 1)The effect of deletion. 2)The performance of the model after deletion. 3)The time cost for deletion. For the effect of deletion, we adopted membership inference attack. Figure\ref{fig:framework} is the framework diagram.

In this section we will introduce the optimization objective of our algorithm. The output distribution of model is different for seen and unseen data. Specifically, for the training data of the model, the confidence that data is classified into a specific category is large, while for the non-training data of the model, the confidence is relatively low. This was also mentioned in \cite{shokri2017membership}. Based this phenomenon, we  consider transforming seen data  into unseen data for the model as our optimization objective. In other word, given data $D$ and model $M$, if a deletion approach could make the models' output distribution about $D$ is similar to third-party data(the same distribution as $D$ but never trained by $M$), we think it could delete $D$ from $M$ successfully.

% \textcolor{red}{as shown in figure 1, we're purchase2 (reference experiments are part of the data set) do the experiment on the binary classification data, The probability density function curves of the output of the model for training data and non-training data do not overlap completely, which means that the output of the model for members of the training set and non-members of the training set is indeed different.}

Let the trained model is $M_{init}$,  $M_{init}$ is trained on dataset $D_t$, data needs to be deleted from $M$ is $D_f$, the rest data is $D_r$, $D_f\cup D_r = D_t$, data detletion algoritm is $\mathcal{U}$, third-party data is $D_{nonmember}$, $D_{nonmember} \cap D_t = \phi$, and $D_{nonmember}$ has the same distribution with $D_t$, $dis$ is distance measurement , like KL divergence, JS divergence, Earth Mover's Distance, etc.

we suppose that deletion algorithm $\mathcal{U}$ removed $D_f$ from $M_{init}$,  the parameters of $M_{init}$  would change after deletion, we use $\Tilde{M}$ to represent the new model.
\begin{equation}
    \mathcal{U}(M_{init}, D_f) = \Tilde{M}
    \label{eq:1}
\end{equation}
Deletion algorithm $\mathcal{U}$ could remove $D_f$ from $M_{init}$ successfully if equation\ref{eq:2} holds:
\begin{equation}
    dis(P(\Tilde{M}(D_{nonmember}), P(\Tilde{M}(D_f))) = 0
    \label{eq:2}
\end{equation}
In light of $D_{nonmember}$ is third-party data both for $M_{init}$ and $\Tilde{M}$, $M_{init}$ and $\Tilde{M}$ have same architecture(like resnet18), thus the following equation should holds:
\begin{equation}
    P(\Tilde{M}(D_{nonmember})) \sim P(M_{init}(D_{nonmember}))
    \label{eq:3}
\end{equation}
Because $\Tilde{M}$ in \ref{eq:2} is agnostic before the end of deletion, therefore we leverage the transformation in \ref{eq:3} to approximately delete $D_f$ from $M_{init}$, we have
\begin{equation}
    dis(P(M_{init}(D_{nonmember})),P(\Tilde{M}(D_f))) = 0
    \label{eq:4}
\end{equation}
Equation\ref{eq:4} means that deleting $D_{f}$ from $M_{init}$ is equivalent to make $P(M_{init}(D_{nonmember}))$ and $P(\Tilde{M}(D_f))$ similar. Equation\ref{eq:4} is our final optimization objective. In next section, we will gave a detailed explain how to optimize this goal. 

\subsection{WGAN Unlearning}
Based on the aforementioned optimization objective, we proposed \textbf{WGAN Unlearning}, whose core idea is training $generator$ and $discriminator$ alternatively until $discriminator$ can not distinguish the difference of output between $D_f$ and $D_{nonmember}$. Specifically, as shown in figure\ref{fig:framework}, the workflow of \textbf{WGAN Unlearning} include two stage, stage one is responsible for deleting $D_f$ from $M_{init}$, while stage two is responsible for verifying the completeness of deletion. In stage one, we initialize $generator$ as $M_{init}$, note that initializing $generator$ as $M_{init}$ means the architecture and parameters of $generator$ are the same as $M_{init}$. During the whole training process, we need to freeze all parameters of $M_{init}$. That is to say, $M_{init}$ only do forward propagation to calculate posterior, and the parameters of $M_{init}$ keep the same all time. Extra third-party data $D_{nonmember}$ is feed into $M_{init}$ to obtain nonmember posterior $P_1$, while $D_{f}$ is feed into $generator$ to obtain posterior $P_2$. In classification task, $P_1$ and $P_2$ are vector, while they are scalar in regression task. In this paper, we focus on classification scenario. After obtaining $P_1$ and $P_2$, we utilize $Sort$ function to sort $P_1$ and $P_2$ in descending order(or ascending order), we will gave a detailed explanation about why need sort them in following section. Here, comparing with standard GAN\cite{goodfellow2014generative}, $S(P_1)$ is equivalent to $real\ data$ and $S(P_2)$ is equivalent to $fake\ data$. we use $S(P_1)$ and $S(P_2)$ to train $generator$ and $discriminator$ alternatively, the same training technique as in standard GAN. Specifically, when training $generator$, freezing the parameters of $discriminator$, while training $discriminator$, freezing the parameters of $generator$. When the distance between $S_P(1)$ and $S_P(2)$ is small enough, we terminate training process. In other words, when $discriminator$ can not distinguish $D_f$ is training data or not of $generator$, the information of $D_f$ is removed successfully from $generator$. 

In stage two, we leverage \textbf{membership inference attack}($MIA$) to test the completeness of deletion for our approach. \textbf{Membership inference attack} is used to determine whether a sample comes from the training data of a trained ML model or not, it was first proposed by \cite{shokri2017membership}, they trained a shadow models for each class, then they utilized these shadow models to train attack model(binary classifier). After finishing training, the attack model could be capable of distinguishing whether the target model trained a sample according to the output of the target model or not. Based on \cite{gulrajani2017improved}, \cite{salem2019ml} further study using one shadow model for all classes, we adapt their method in this paper. Specifically, for simplicity, we first initialize the shadow model as the same architecture as $M_{init}$, and train the shadow model. Then, we train the attack model according to the output of training data and non-training data of the shadow model. Next, we feed $D_f$ into $generator$ obtaining from stage one. If the attack model still infers $D_f$ as training data of $generator$, then our approach failed to delete $D_f$ from $generator$ in stage one. Otherwise, our method can remove $D_f$ from $generator$ successfully. 

\subsection{From GAN to WGAN GP}
Since \cite{goodfellow2014generative} first proposed GAN, many studies of GAN have appeared. \cite{arjovsky2017towards} Point out that there exits gradient vanishing and mode collapse in original GAN. They also analyzed the reasons for these problems. Based on \cite{goodfellow2014generative}, \cite{arjovsky2017wasserstein} replaced the loss function of discriminator with Earth Mover distance, and proposed WGAN. WGAN alleviated the problem of gradient vanishing and model collapse, but they simply clip parameters of discriminator to satisfy Lipschitz condition, which slow down the whole training process of GAN. In order to address this problem, \cite{gulrajani2017improved} proposed WGAN with gradient penalty(WGAN GP), which add an extra term of gradient penalty on the loss function of discriminator instead of clipping parameters. WGAN GP could efficiently accelerate the training process. In light of the advantages of WGAN GP, we adapt WGAN GP as our base algorithm.  

Based on WGAN GP, we redesign loss functions both for $discriminator$ and $generator$. We use $L(D)$ and $L(G)$ to represent loss funciton of $discriminator$ and $generator$ respectively:
\begin{align}
    L_{D} &=  -\mathbb{E}_{x \sim \mathbb{P}_{nonmember}}[D(S(M_{init}(x)))] \notag \\ &+ \mathbb{E}_{z \sim \mathbb{P}_{f}}[D(S(G(z)))] \notag \\ &+ \lambda\mathbb{E}_{\hat{x} \sim \mathbb{\hat{P}}}[||\nabla_{\hat{x}}||_p - 1]^2
    \label{eq:5}
\end{align}

\begin{align}
    L_{G} &= \alpha(- \mathbb{E}_{z \sim \mathbb{P}_{f}}[D(S(G(z)))]) \notag \\ 
    &+ (1-\alpha) \mathop{L}\limits_{(x,y)\in D_{r}}(G(x),y)
    \label{eq:6}
\end{align}
Wherein, $\hat{x}=\epsilon S(P_1) + (1-\epsilon)S(P_2)$, $\epsilon \sim Uniform(0, 1)$, this means that $\hat{x}$ is obtained by uniform sampling on the connecting line of two probability distributions $S(P_1)$ and $S_(P_2)$. Moreover, we add an additional term $\mathop{L}_{(x,y)\in D_r}(G(x),y)$ into the loss function of $generator$ to keep the performance of $generator$, $L$ is loss function of training $M_{init}$, like CrossEntropy in classification task. The performance degradation will be high if without this term in our experiment. We use hyperparameter $\alpha$ to seek a balance between deletion and performance. Note that there is no need for all $D_r$, just a small subset of $D_r$ is enough(like 1/100). $S$ is $sort$ function, and we will give a detailed explanation in the following section.

\subsection{Sort}
\begin{figure}[htbp]
    \centering
    \includegraphics[scale=0.25]{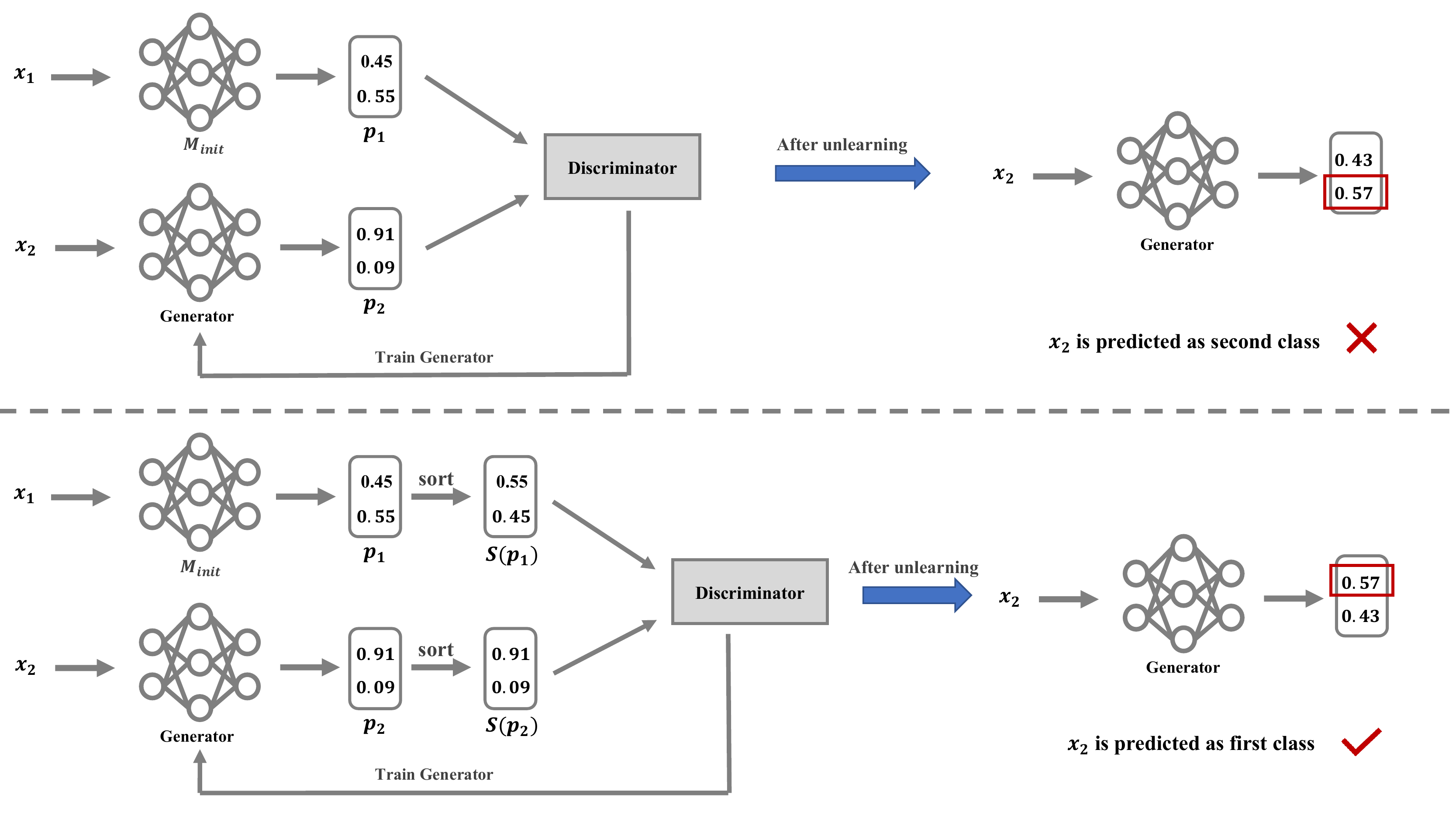}
    \caption{No sort vs sort.}
    \label{fig:sort}
\end{figure}

Now we explain why we need the $Sort$ function. We want the model $\Tilde{M}$ to be able to classify $D_f$ correctly. Intuitively, though $D_f$ is removed from $\Tilde{M}$, a well trained model $\Tilde{M}$ should have the generalization capability to perform well on $D_f$. Which is like validation or test data, the model never trained on them, but the model could classify correctly for most of them. Specifically, as shown in figure\ref{fig:sort}, assuming that we are handling with binary classification task, $M$ is a well trained ML model, the label of $x_1$ and $x_2$ is category 1 and category 2 respectively, and the label is encoded as one-hot vector. $x_1$ is not trained by $M$, and the corresponding output of $M$ is $P_1 = [0.45,0.55]$, $x_2$ is trained by $M$, the corresponding output of $M$ is $P_2 = [0.91, 0.09]$, that is to say, $M$ classify $x_1$ and $x_2$ into category 2 and category 1 respectively correctly, cause predictions of $M$ are consistent with data label. The difference between $P_1$ and $P_2$ is that the class confidence of $P_2$(0.91) is large as $x_2$ has been trained by $M$ while $P_1$(0.55) is relative small as $x_1$ have not been trained by $M$. We leverage non-training data $x_1$ to removed $x_2$ from $M$. Assuming $P_2=[0.43,0.57]$ after removing $x_2$ from $M$ without $sort$, $M$ has almost erased $x_2$. However, $M$ misclassify $x_2$ as category 2 after deleting through GAN, we desired result is $P_2=[0.57,0.43]$. Therefore, the intention of introducing $sort$ function is making the model could still classify the data correctly after removing the data from the model. Note that, we can also achieve the above goal without $sort$ if the label of non-training data $x_1$ is the same to $x_2$, but in a real scenario, user data may contain multiple classes, it may difficult to find a non-training data that contains all classes of user data. 

\subsection{Membership inference attack}
The aim of membership inference attack(MIA) is to determine whether a sample is used to train a model or not. The basic attack idea of MIA is training the shadow model to imitate the target model, then the attacker can attack the target model indirectly by attacking the shadow model. MIA was first proposed by \cite{shokri2017membership}, they trained a shadow model for each class of data, then they utilized these trained shadow models to train attack model(binary classifier). After finishing training, the attack model could be capable of distinguishing whether the target model trained a sample or not according to the output of the target model. Based on \cite{shokri2017membership}, \cite{salem2019ml} further studied using one shadow model for all classes, they proposed three type of adversary that gradually improved the attacker's capabilities, we use the first adversary as our attacker. In our attack scenario, the distribution of training data of target model is available  for the attacker, and for convenience, we initialied the architecture of shadow model as same as target model. Specifically, we divide the original dataset of $D_s$ into two disjoint datasets, $D_s$ and $D_t$. $D_t$ is used to train the target model $M_{init}$. $D_f \in D_{t}$, and $D_f$ need to be forgotten in $M_{init}$. we further separate $D_s$ into $D_{in}$ and $D_{out}$, then train the shadow model $M_{shadow}$ on $D_{in}$. The output of the shadow model about $D_{in}$ and $D_{out}$ are $P_{in}$ and $P_{out}$. We label $P_{in}$ and $P_{out}$ with training and non-training category to form a new dataset, this new dataset is used to train the attack model(usually is a binary classifier). A well trained attack model is capable to determine whether a sample was trained by the target model $M_{init}$ or not according to the output of $M_{init}$. If the attacker inference $D_f$ as non-training data of $generator$, then the information about $D_f$ in $generator$ was successfully removed through our approach.

\subsection{Retrain from scratch}
Retraining the model on the remaining dataset,  which we called retrain from scratch, is prohibitively expensive. One the one hand, training a deep classifier on big dataset, like ImageNet, would take few days and cost huge computational resources. One the other hand, the service provider has limited storage to store all user data,  they might delete some of the data after the data was trained by the model, in this scenario, retraining the model requires recollecting user's data,  but user may not be willing to do that. Though retraining the model from scratch is unfeasible in many situation, it performs well in some aspects. For example, retraining the model could comletely delete data, because the data was never involved in retraining process, therefore the retrained model naturally exclude the data.  We use retrain as baseline in this paper.

\section{Evaluation}
\subsection{Evaluation metrics}
We evaluated our method from three aspects, that is, effectiveness, permformance and time. Effectiveness indicates that whether the data was completely removed from the model. Performance demonstrates the accuracy of model on test dataset after deletion. Time shows that how long would take to delete. We utilize FNR, Test Accuracy and Time Cost to represent the above three aspects respectively.

\paragraph{FNR} We adapt the results of member inference attack to calculate FNR(False Negative Rate). The definition of FNR is given by the following equation:
\begin{equation}
    FNR = \frac{FN}{TP+FN}
\end{equation}
wherein, TP means that MIA inference training data as training data of model $M$; FP means that MIA inference non-training data as training data of model $M$; TN means that MIA inference non-training data as non-training data of model $M$; FN means that MIA inference training data as non-training data of model $M$.When finishing deletion,  the data would behave like the non-training data for model. If the data imformation is completely removed from the model, MIA attack model would treat the training data as non-training data of model, thus FN will be large, TP will be small, the corresponding FNR will be large;else the FNR will be small.

\paragraph{Test Accuracy}
Deleting data from a trained model would do harm to the performance of model, which reflect on accuracy decrease on test data. Retrain the model from scratch would also has a negative impact on test accuracy. The key point is how to reduce the performance gap between original model and Unlearned model. 

\paragraph{Time Cost}
Time cost for deletion is also a key indicator to evaluate the efficiency of unlearning method. Retrain as a baseline method, one of its disadvantages is that the time cost is too much. Note that, Time cost is not consistent  in different scenarios. For example, retraining a deep classifier on MNIST would take a few minutes while it would cost a few days on ImageNet, since the  instance of sample in ImageNet is bigger than that in MNIST. Therefore, for fair comparison, we evaluate our method on different settings and different datasets. 

\subsection{Datasets}
We evaluated our approach on Fashion, SVHN, CIFAR10, CIFAR100 and Purchase, the following is a brief introduction about these datasets.

\paragraph{Fashion} The Fashion\cite{xiao2017fashion} dataset contains 70,000 different samples of products from 10 categories, such as T-shirts, pants, skirts, etc. The training dataset includes 60,000 samples, and the testing dataset includes 10,000 samples. Each sample is a grayscale image of $28\times28$.

\paragraph{SVHN} The Street View House Numbers (SVHN)\cite{netzer2011reading} Dataset comes from house numbers in Google Street View images, which contains 73257  samples of training data and 26032 of testing data, each sample is $32 \times 32 $ RGB digit number. SVHN has two different format, one is the image with character level bounding boxes, another is the image centered around a single character, we use the latter in our experiment.

\paragraph{CIFAR} CIFAR\cite{CIFAR} dataset is obtained from real-world, for example, pictures of airplane, bird, cat, etc. It consists of 60000 RGB images with the size of $32 \times 32$, wherein 5000 for training data, 10000 for testing data. There are two versions of CIFAR, CIFAR10 and CIFAR100. In CIFAR10, it has 10 classes, with 6000 samples per class. While in CIFAR100, it contains 100 classes including 600 images each.

\paragraph{Purchase} Purchase\cite{Purchase} is an unlabeled dataset from Kaggle's "acquire valued shoppers" competition. The purpose of the competition is to design a coupon recommendation strategy. Each data sample contains the transaction record of the user over a year, such as product name, quality, date, etc. Analogous to \cite{shokri2017membership}, we use a simplified version containing 19,324 samples, each containing 600 features. At the same time, we leverage clustering algorithm Kmeans to cluster them into 10, 20, 50, and 100 categories to form four datasets, and we represent them with Purchase10, Purchase20, Purchase50, and Purchase100, respectively.

\subsection{Architecture of model}
Because the complexity of the dataset is different, thus we use different architectures of model on different datasets. The details can be found in table\ref{target_model}.

\begin{table}[h]
\centering
\tiny
\scalebox{2}{
\begin{tabular}{|c|c|}
\hline
\textbf{Datasets} & \textbf{Models}       \\ \hline
FASHION                       & LeNet\cite{lecun1995learning}      \\ \hline
SVHN                          & 2$\times$Conv+2$\times$FC \\ \hline
CIFAR10                       & Resnet18\cite{he2016deep}    \\ \hline
CIFAR100                      & Resnet18    \\ \hline
Purchase[10,20,50,100]    & 3$\times$FC        \\ \hline
\end{tabular}
}
\caption{Architectures for different datasets}
\label{target_model}
\end{table}

Since overfitting on training data is the key to implement membership inference attack[], thus as shown in table\ref{overfit}, we overfit the model on each corresponding dataset.

\begin{table}[h]
\centering
\small
\scalebox{1}{
\begin{tabular}{c|cc}
\textbf{Datasets}    & \textbf{Training accuracy} & \textbf{Testing accuracy} \\ \hline
FASHION     & 0.98              & 0.846            \\ 
SVHN        & 0.993             & 0.829            \\ 
CIFAR10     & 0.997             & 0.536            \\ 
CIFAR100    & 0.98              & 0.319            \\ 
Purchase10  & 0.99              & 0.811            \\ 
Purchase20  & 1                 & 0.751            \\ 
Purchase50  & 1                 & 0.693            \\ 
Purchase100 & 1                 & 0.617            \\ 
\end{tabular}
}
\caption{Overfit level on different datasets}
\label{overfit}
\end{table}

\subsection{Experiment results}
In this section, we give a detailed introduction of our experiment. In our experiment, the data $D_{f}$ needs to be deleted and the third-party data $D_{nonmember}$ is consist of 500 randomly sampled from the training set $D$ and $D_{test}$ respectively, the rest of $D_{test}$ forms a new test set.

\paragraph{1)Effectiveness of deletion}:
We have implemented MIA both on original model $M_{init}$ and unlearned model $\tilde{M}$. MIA model aim to determine whether $D_f$ is trained by $M_{init}$($\tilde{M}$) or not. In ideal case, MIA model would distinguish $D_f$ and $D_{nonmember}$ as training data and non-training data of $M_{init}$ before deletion, MIA model would distinguish both $D_f$ and $D_{nonmember}$ as non-training data of $\tilde{M}$ after deletion, since MIA attack model found no difference of the output of  $\tilde{M}$ between $D_f$ and $D_{nonmember}$.

\begin{figure*}[htbp]
\centering
\subfloat[Original model]{
\begin{minipage}[t]{0.3\textwidth}
\centering
\includegraphics[width=5.5cm]{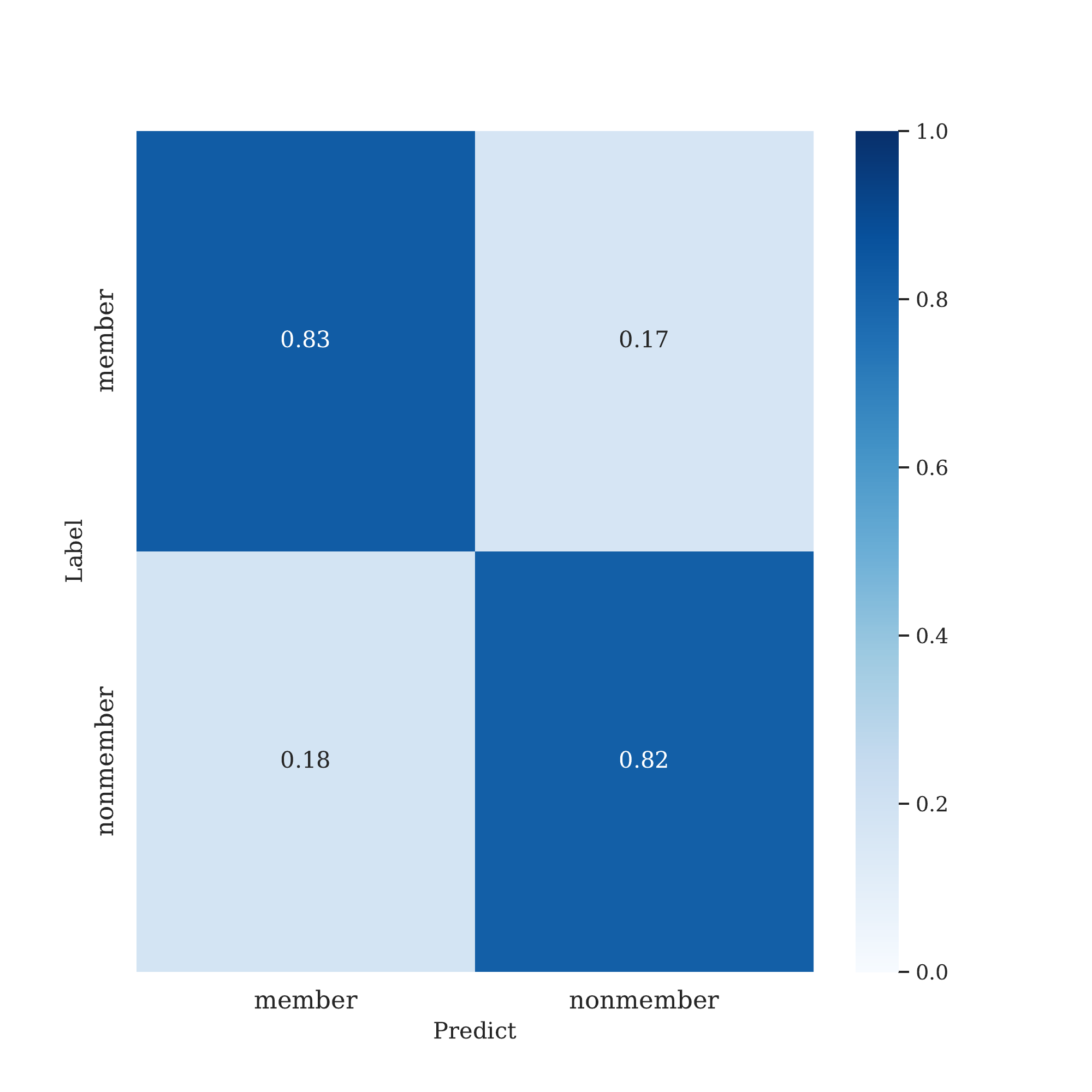}
% \caption{Target model}
\label{CF_Target_CIFAR100}
\end{minipage}
}
\subfloat[Retrain model]{
\begin{minipage}[t]{0.3\textwidth}
\centering
\includegraphics[width=5.5cm]{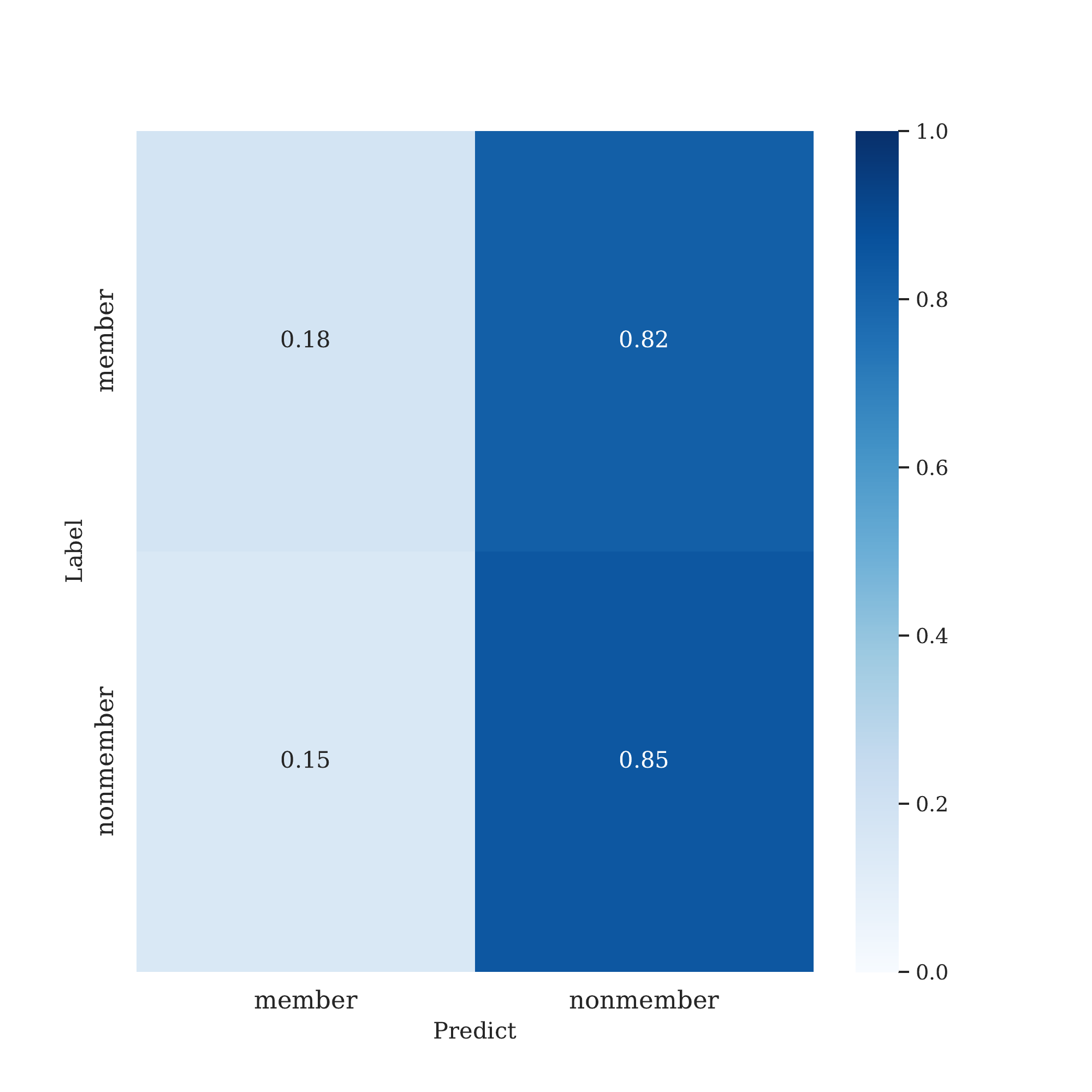}
% \caption{Retrain}
\label{CF_Retrain}
\end{minipage}
}
\subfloat[Unlearn model]{
\begin{minipage}[t]{0.3\textwidth}
\centering
\includegraphics[width=5.5cm]{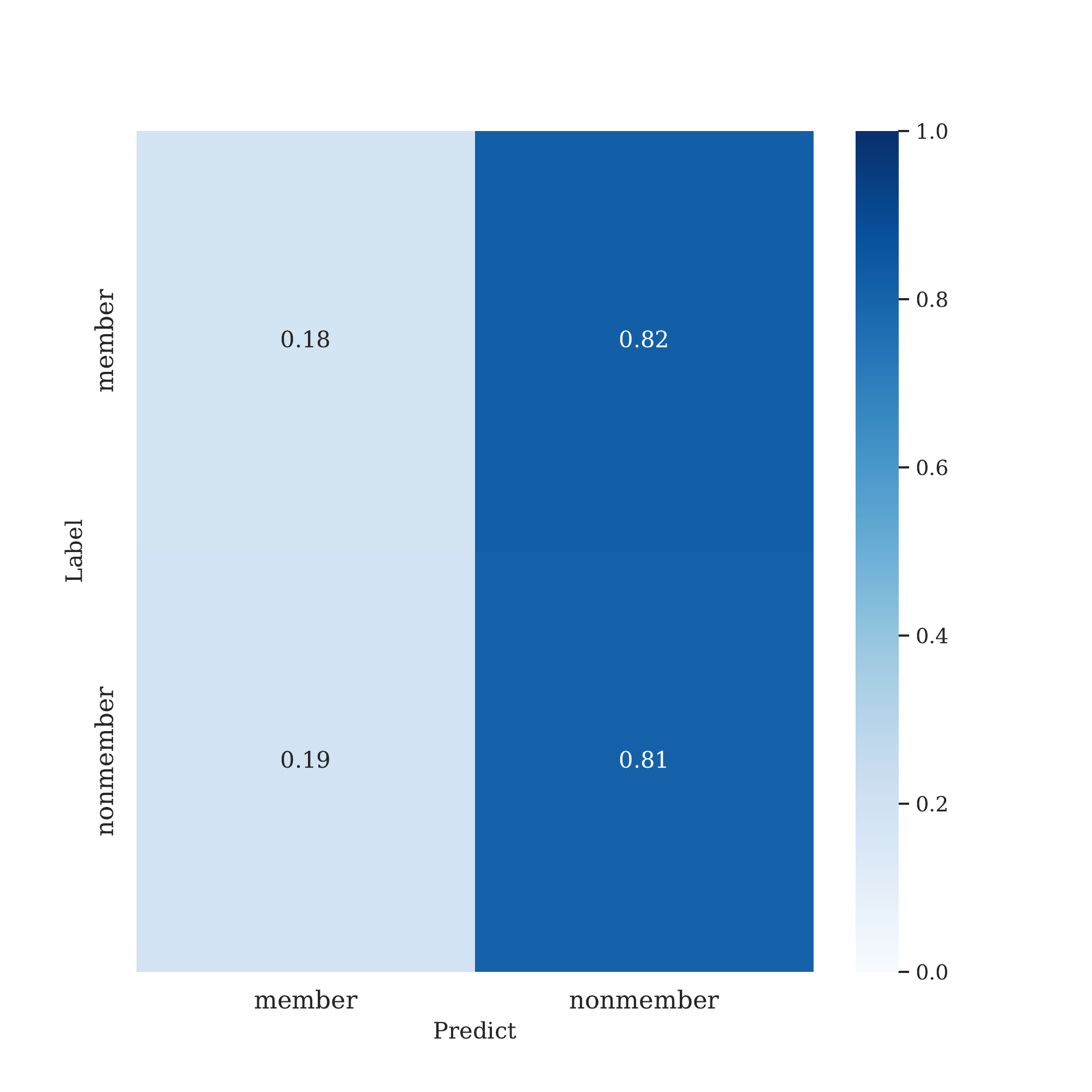}
% \caption{Unlearn}
\label{CF_Unlearn}
\end{minipage}
}
\caption{Confusion matrix of MIA on CIFAR100.}  
\label{CM_CIFAR100}
\end{figure*}

\begin{figure*}[h]
\centering

\subfloat[Original model]{
\begin{minipage}[t]{0.3\textwidth}
\centering
\includegraphics[width=5.5cm]{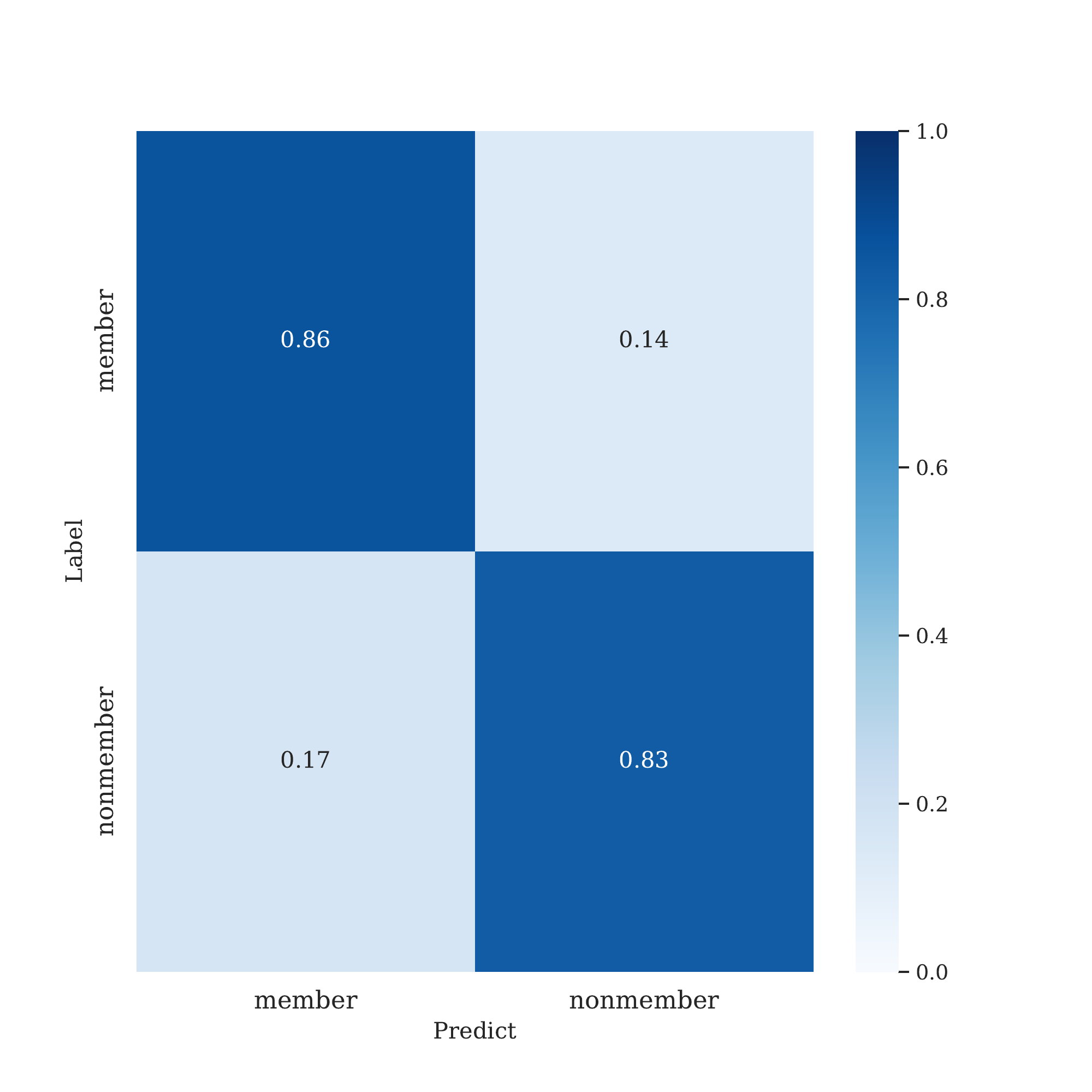}
% \caption{Target model}
\label{CF_Target}
\end{minipage}
}
\subfloat[Retrain model]{
\begin{minipage}[t]{0.3\textwidth}
\centering
\includegraphics[width=5.5cm]{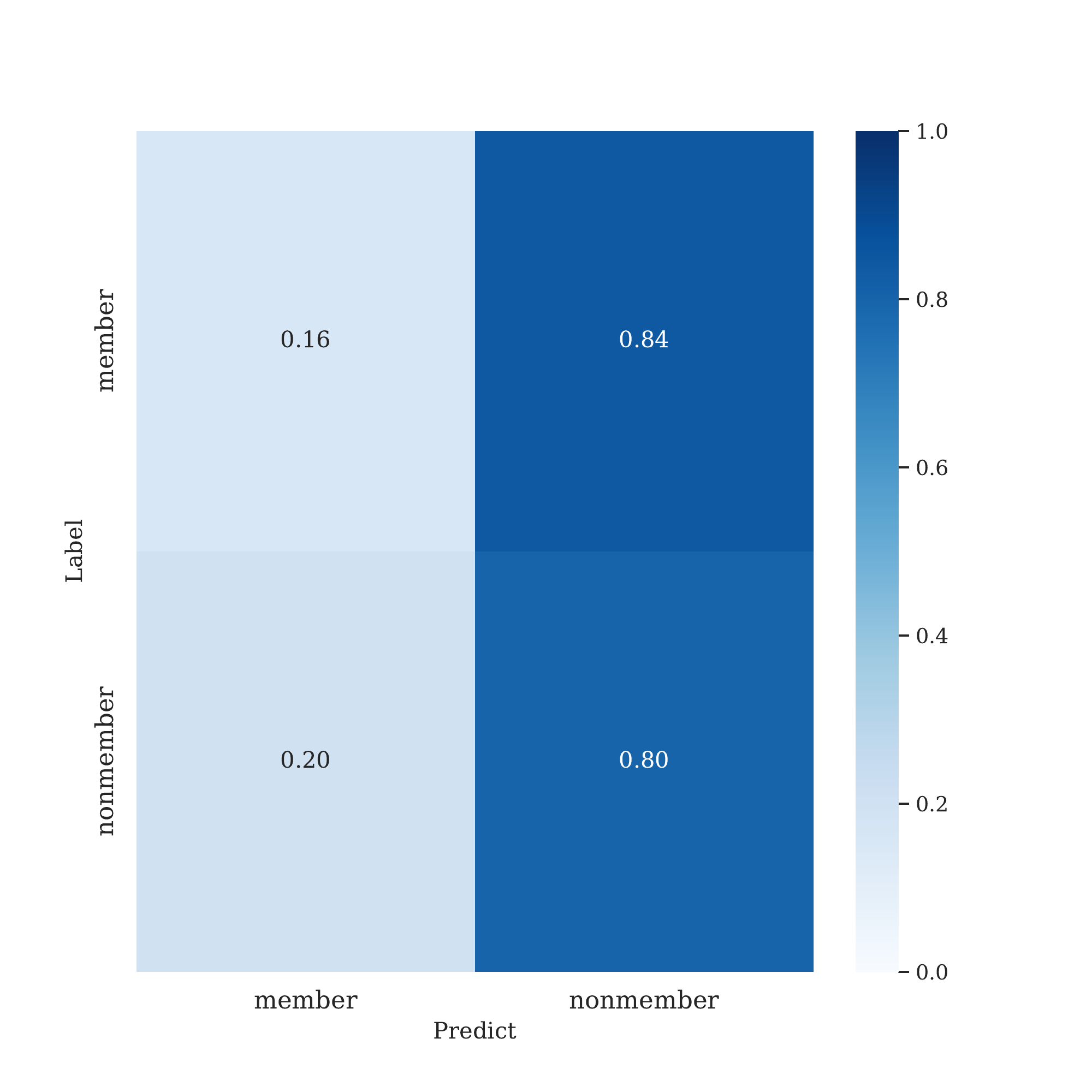}
% \caption{Retrain}
\label{CF_Retrain}
\end{minipage}
}
\subfloat[Unlearn model]{
\begin{minipage}[t]{0.3\textwidth}
\centering
\includegraphics[width=5.5cm]{ 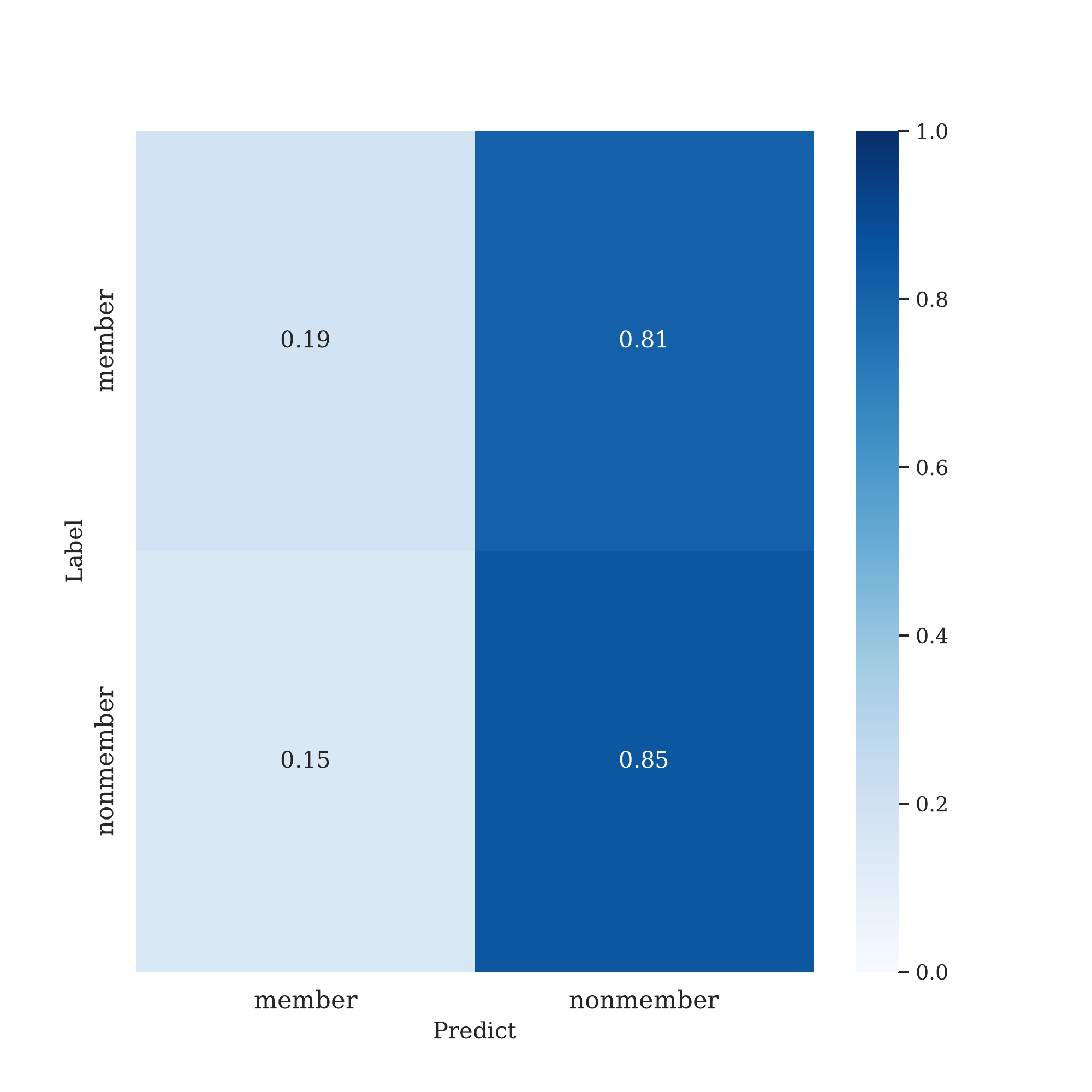}
% \caption{Unlearn}
\label{CF_Unlearn}
\end{minipage}
}
\caption{Confusion matrix of MIA on Purchase100.}
\label{CM_Purchase100}
\end{figure*}

As shown in Fig\ref{CM_CIFAR100}, the TP and TN of MIA are 0.83, 0.82 respectively for original model $M_{init}$, which indicates that MIA is capable to distinguish most of $D_f$ and $D_{nonmember}$. For retain model, the TP of	MIA decrease to 0.18 while FN increase to 0.82, it means that most of $D_f$ are determined as non-training data of $\tilde{M}$ by MIA, this demonstrates that the deleting effectiveness of retrain is excellent. For our method, the TP, FN and TN are 0.18, 0.82 and 0.81 respectively, which are similar with retrain. Fig\ref{CM_Purchase100} is experiment on Purchase100, the conclusion is analogous to CIFAR100. 

\begin{figure}[h]
    \centering
    \includegraphics[scale=0.35]{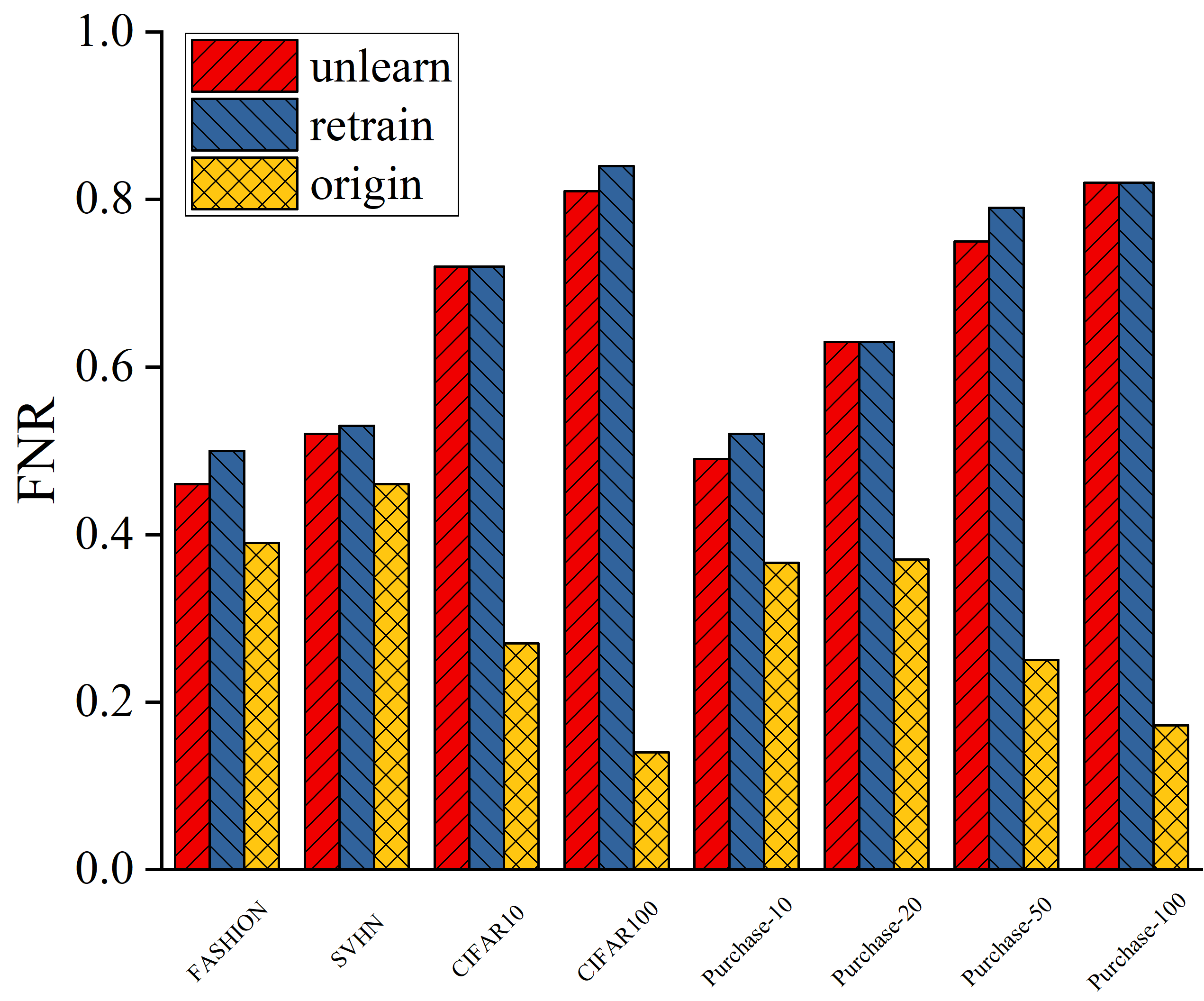}
    \caption{FNR of membership inference attack.}
    \label{FNR}
\end{figure}

The results of FNR are shown in Fig\ref{FNR}. The FNR of the original model before deletion is relatively low on each dataset, which means that the MIA model classified correctly as non-training data for most of $D_{f}$. While the FNR of MIA on the unlearned model is larger than the original model for all datasets, these results indicate that our method could indeed remove the information of $D_f$ from a trained model. Meanwhile, for CIFAR10, Purchase20, and Purchase100, our approach is almost the same as retrain, while it is slightly lower on others datasets compared to retrain.

% % =============================confusion matrix================

\begin{figure}[h]
    \centering
    \includegraphics[scale=0.35]{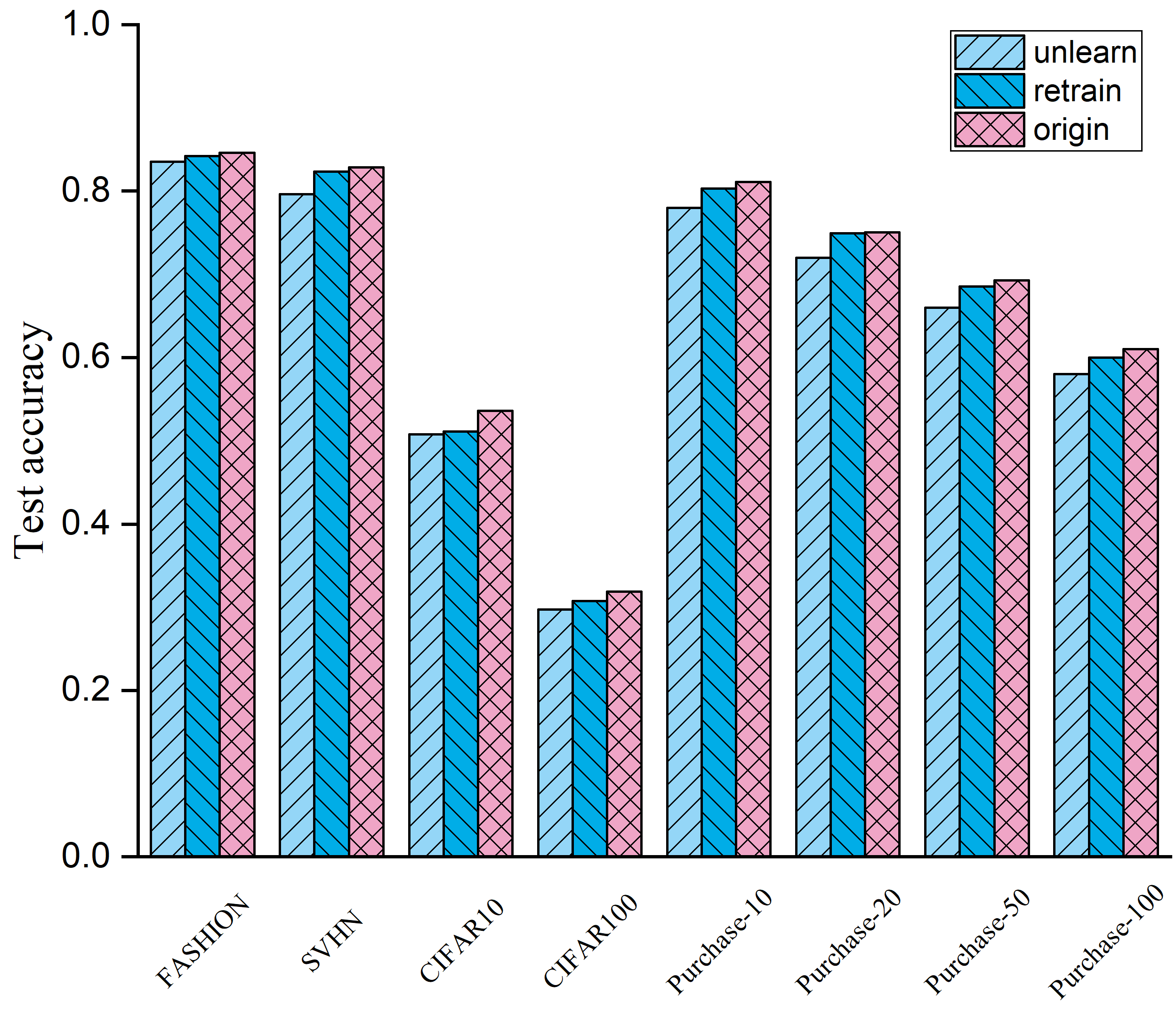}
    \caption{Model's accuracy on test data.}
    \label{test_acc}
\end{figure}

% % =============================confusion matrix================
\paragraph{2)Accuracy on test data}:
We use the accuracy on test data to measure the impact of deleting method. As shown in Fig\ref{test_acc}, the accuracy of original model is higher both than retrain model and unlearn model on all datasets, this is reasonable, cause most of deletion method have negative impact to model performance more or less. Moreover, the accuracy of our method is close to retrain on FASHION, CIFAR10 and CIFAR100, while that is slightly lower compared to retrain on other datasets, but the gap is in acceptable range.

\paragraph{3)Time Cost for deletion}:
As mentioned in the previous chapter, time cost for unlearning is also the core indicator to evaluate deletion algorithm, the shorter time for deletion, the better performance of algorithm. Note that, as the number of training data and parameter of model increase, the time cost for retraining is also increasing, \cite{bourtoule2019machine} also mentioned this problem. Therefore, to evaluate fairly, we experimented both on simple and complex scenarios.  we randomly sampled three disjoint datasets from the training data of CIFAR10, the number of instance for each are 5000, 10000 and 25000. Then, these three datasets are used to train resnet18, resnet50 and resnet101 respectively. The parameters of each model are different, among them, resnet18 is the smallest, which is 1.2 million, and resnet101 is the largest, which is 4.5 million. The trained model is treated as the original model. $D_f$ are randomly sampled from each training data of original model. Deletion method aims to remove $D_{f}$ from each original model. 

\begin{figure}[hbtp]
    \centering
    \includegraphics[scale=0.15]{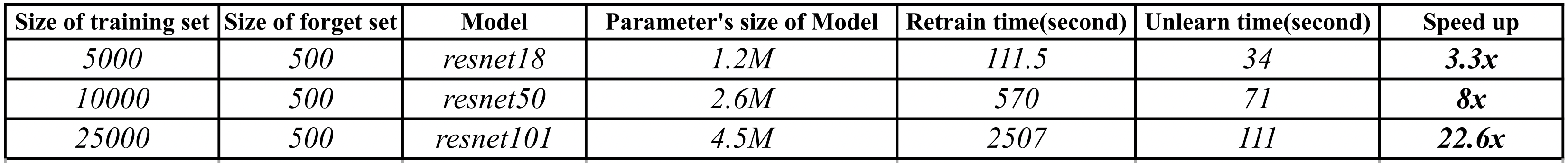}
    \caption{Time cost of our approch on CIFAR10.}
    \label{time}
\end{figure}

As shown in table\ref{time}, we can see that our approach is better than retrain both in three scenarios. In simple scenarios, our algorithm is 3.3 times faster than retrain while that is 22.6 times faster than retrain in complex scenario. Although the time cost of our algorithm increase as the scenario become more complex, our increase rate is much smaller than retrain. From this point, our algorithm has a great advantage over retrain. 

\begin{figure}[hbtp]
    \centering
    \includegraphics[scale=0.35]{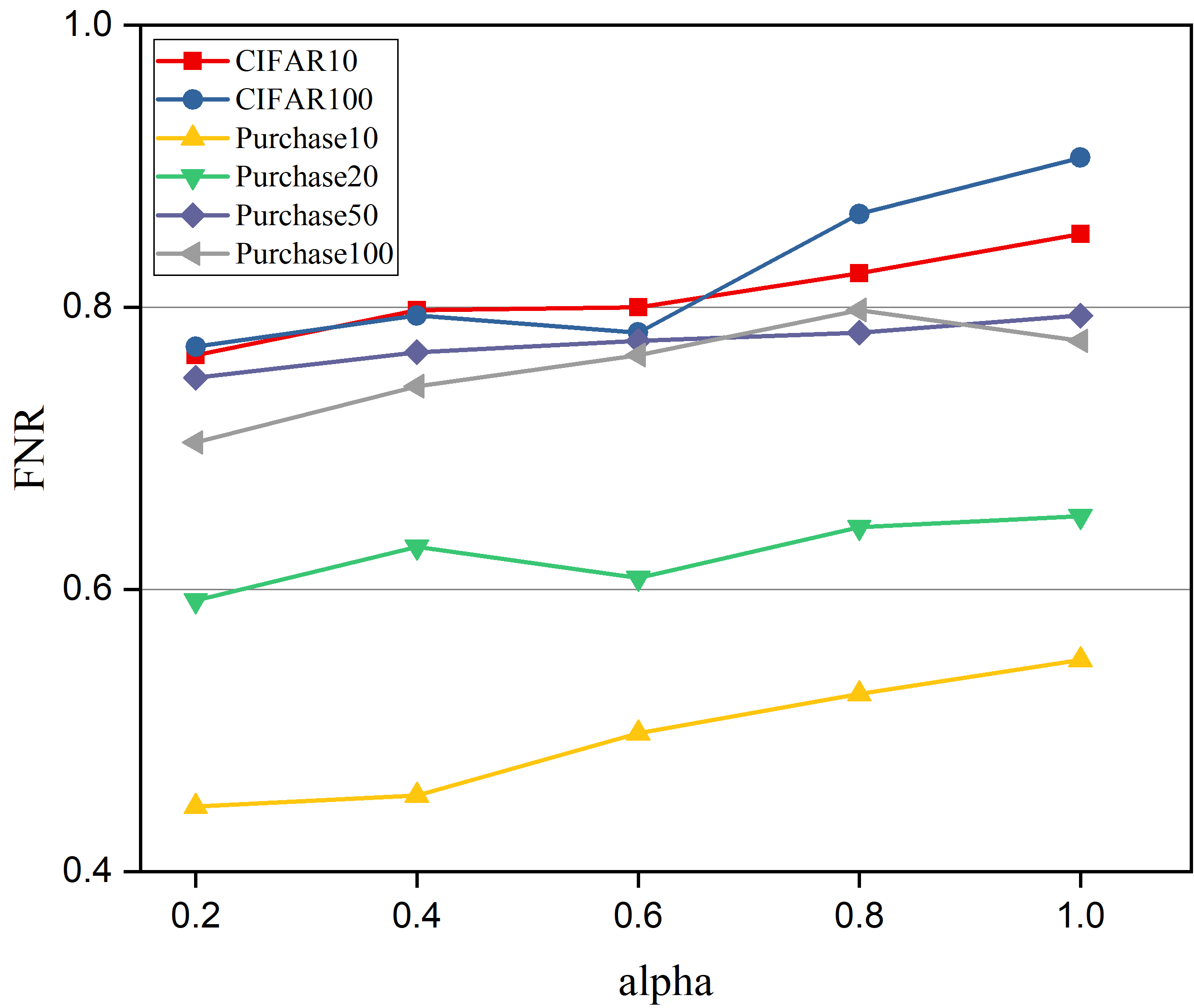}
    \caption{Impact of FNR.}
    \label{alpha_FNR}
\end{figure}

\begin{figure}[hbtp]
    \centering
    \includegraphics[scale=0.35]{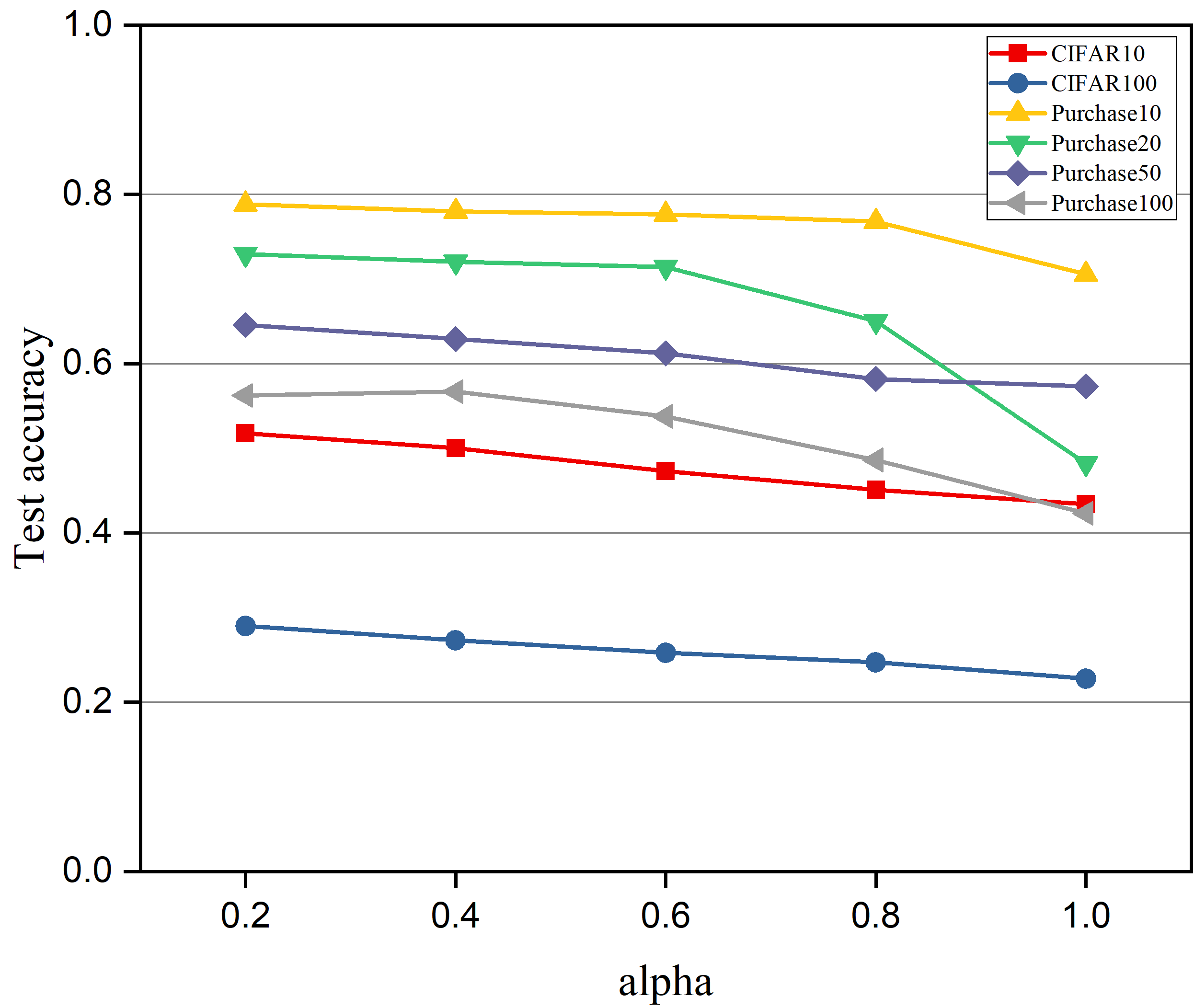}
    \caption{Impact of performance.}
    \label{alpha_testacc}
\end{figure}

\paragraph{4)Impact of alpha}
We found that different values of alpha will have different effects. We take 5 values of alpha from 0.2 to 1.0 at equal intervals to evaluate the impact of alpha. As shown in Fig\ref{alpha_FNR} \ref{alpha_testacc}, as alpha increases, FNR of MIA is raising, while Test accuracy is decreasing, which indicates that deletion is more complete with higher value of alpha, but the performance is less competitive after deletion. For different datasets and model, the value of alpha is not constant. For example, when increasing alpha from $0.6$ to $0.8$ on Purchase20, the Test accuracy of unlearned model drop from $71\%$ to $64\%$, reduced by $7\%$; and the FNR of MIA raised from $61\%$ to $64\%$, increased by $3\%$, which demonstrates that $7\%$ performance degradation of model brings $3\%$ escalation of deletion. In conclusion, alpha is vary for different datasets, bigger alpha more consider about deletion while smaller alpha more consider performance, thus it is necessary to find a appropriate alpha to balance between performance and deletion.

\begin{figure*}[h]
\centering

\subfloat[Training/testing accuracy(1/0.91)]{
\begin{minipage}[t]{0.3\textwidth}
\centering
\includegraphics[width=5.5cm]{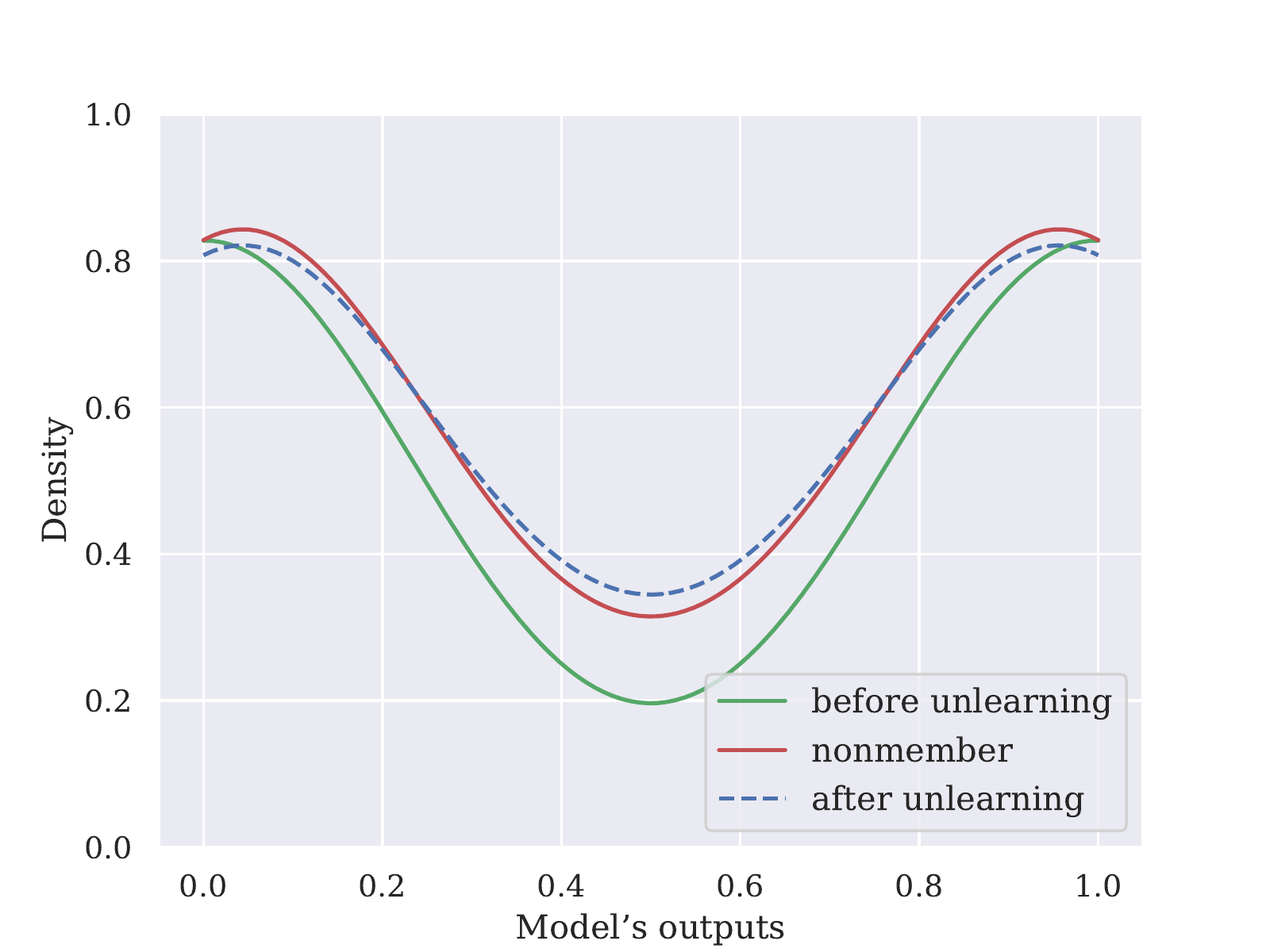}
% \caption{Target model}
\label{Training/testing accuracy(1/0.91)}
\end{minipage}
}
\subfloat[Training/testing accuracy(1/0.86)]{
\begin{minipage}[t]{0.3\textwidth}
\centering
\includegraphics[width=5.5cm]{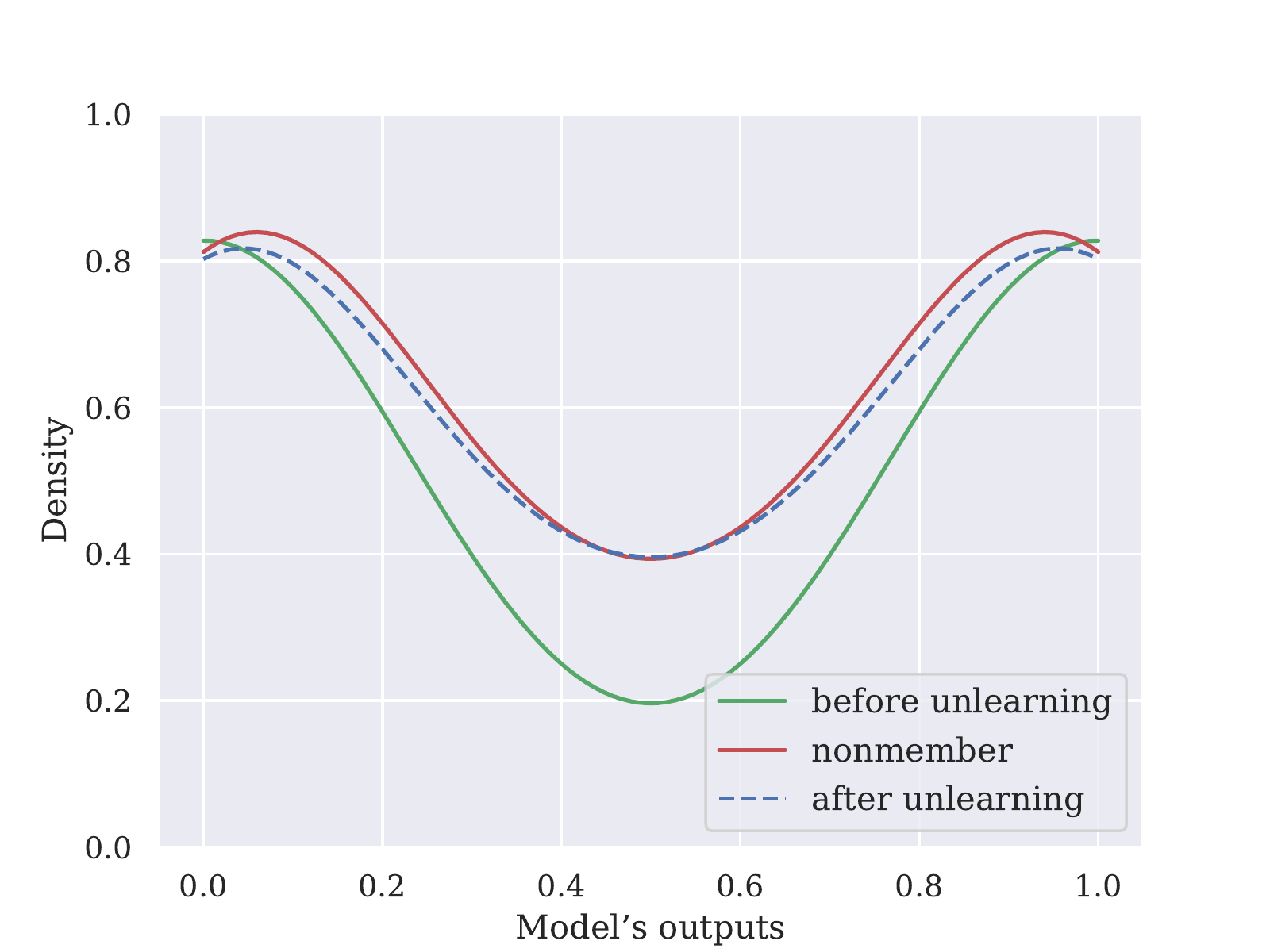}
% \caption{Retrain}
\label{Training/testing accuracy(1/0.86)}
\end{minipage}
}
\subfloat[Training/testing accuracy(0.97/0.78)]{
\begin{minipage}[t]{0.3\textwidth}
\centering
\includegraphics[width=5.5cm]{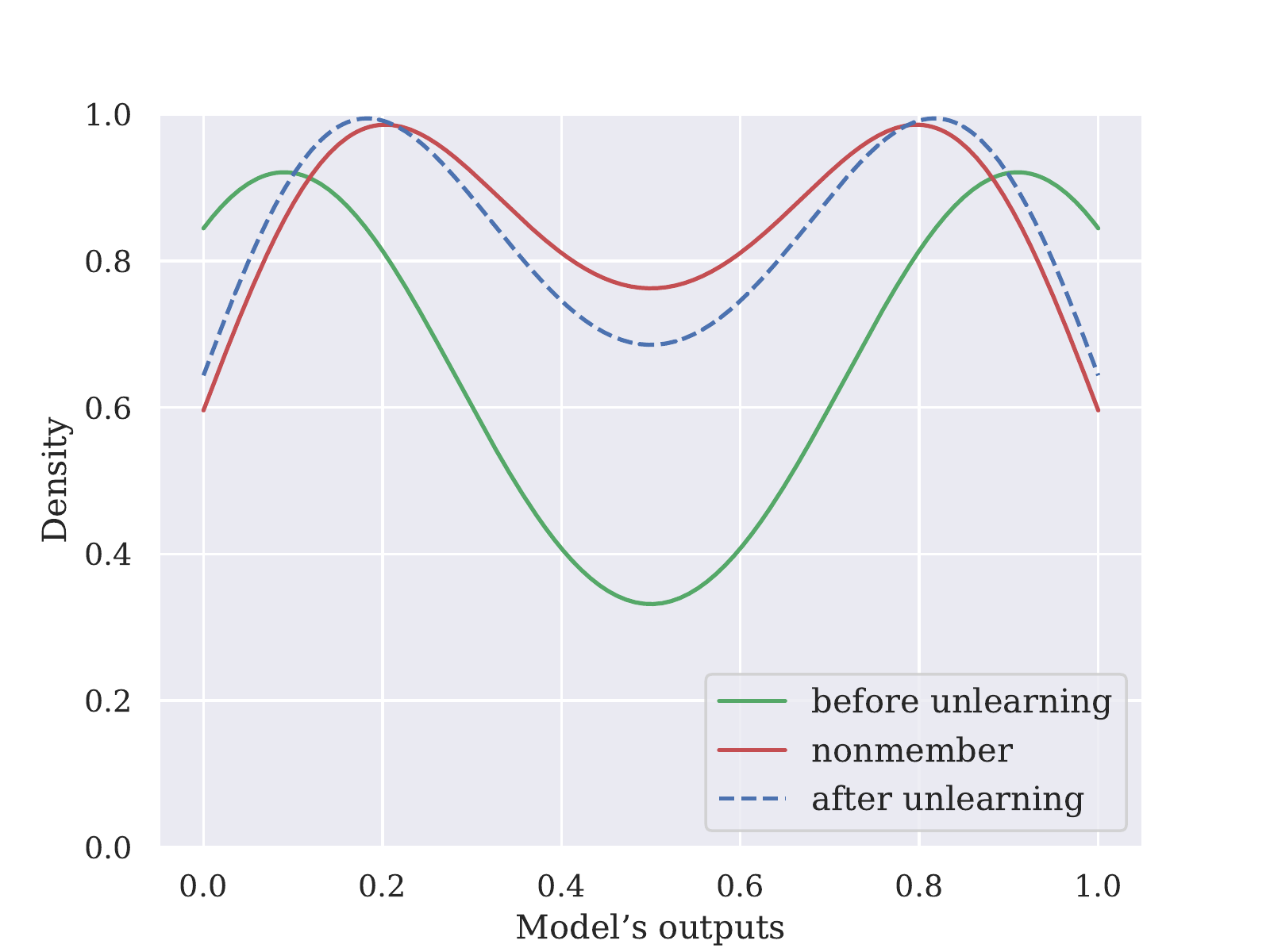}
% \caption{Unlearn}
\label{Training/testing accuracy(0.97/0.78)}
\end{minipage}
}
\caption{overfit.}
\label{diff_overfit}
\end{figure*}

\paragraph{5)Impact of overfit}
We’ve further studied the impact of overfitting. We divided Purchase dataset into two categories by utilizing the clustering algorithm kmeans, and then trained the model on this dataset with varying degrees of overfitting. The model we used containing three fully connected layers, and the activation function is $softmax$, the output dimension of the model is two. Before and after deletion, we feed the data $D_{f}$ and the third-party data $D_{nonmember}$ into the original model respectively, then draw curve of the probability density function(PDF) of corresponding output. Experiment results are shown in Fig\ref{diff_overfit}, the green and red solid line represents the PDF curve of $D_f$ and $D_{nonmember}$ before deletion. These two lines does not fully overlap, which indicates that the model could distinguish between $D_f$ and $D_{nonmember}$. The blue dash line is the PDF curve of unlearned model using our method, it almost overlap with the red solid line, this demonstrates that there is no difference between $D_f$ and $D_{nonmember}$ for the unlearned model. In other words, our approach could removed specific data from a trained model. From (a) to (c), we gradually strengthen the level of overfitting, the shape and location of red solid line and blue dash line are similar both on these three situation, this shows that our method could effectively remove data for different overfitting level of model.

\section{Related works}

The term “Machine unlearning” was first introduced by \cite{cao2015towards}, they proposed statistical query-based unlearning, which could efficiently erase data information from a trained model, but it failed to apply to neural network. \cite{ginart2019making} proposes two deletion algorithms for the clustering algorithm k-means. \cite{bourtoule2021machine} proposed method similar to ensemble learning, which divides the original data set into multiple blocks, and each substitute model trains each blocks independently, so that the information of the data is limited to the corresponding substitute, deletion data only needs to retrain corresponding substitute model, moreover, to accelerate the unlearning process, they further divide the data block into multiple data slices, thus each substitute model incremental training on each slice, and save the intermediate parameters of model at the same time. This method changed the paradigm of machine learning, which may be unapplicable in some scenarios, and it needs storage resource to save the intermediate parameters of model, that is prohibitively expensive for large model. \cite{guo2020certified} proposed a certified removal method related to differential privacy, they applied Newton's method to delete data in the linear model, and added random noise to the loss function of the model to make it indistinguishable between unlearned model and retrained model,  the disadvantage of this method is that the computational complexity is relatively high (main for calculating the inverse of the Hessian matrix), and it is also not suitable for some models where the loss function is non-convex. \cite{golatkar2020eternal} studied adding noise on the model parameters to delete a specific class (or a subset of a specific class) in the classification task, the drawback of this method is also the high computational complexity . \cite{liu2020federated} studied deleting data in federated learning scenario, during the training phase, the central parameter server saves the updated parameters for each round, when deleting data, just retrain the model on remaining data, and retrain phase can be accelerate by load intermediate parameters, the disadvantage of this method is that caching the parameters would consume a lot of storage resource, especially for complex model. \cite{izzo2021approximate} studied the problem data deletion in linear models and logistic regression models, and proposed an approximate deletion method, whose computational cost is linear with data dimension. Recently, \cite{brophy2021machine}studied how to delete data in random forests.

\section{Conclusions}
Experiments results shows the feasibility of GAN-based unleanring, and our method is faster for deleting data from a trained model than retrain, especially for complicated scenarios(like big dataset and complex model). Moreover, our approach is storage-free compared to previous works, there is no needs for storage resource to cache the intermidiate parameters. The drawback of our method is that the performance degradation is slight higher than retrain, which indicates that it would do harm to the model for deleting data by utilizing GAN-based unlearning, how to minimize the degradation is important, we treat this problem as future work.

\small
\bibliographystyle{plain}
\bibliography{ref}

\end{document}